\documentclass{article}
\usepackage{ICASSP2022,amsmath,amssymb,graphicx}
\usepackage{xcolor}
\usepackage{hyperref}
\usepackage{booktabs,cite}
\usepackage[super]{nth}

\title{A two-step approach to leverage contextual data: speech recognition in air-traffic communications}
\name{Iuliia Nigmatulina$^{~\dagger, \ddagger}$, Juan Zuluaga-Gomez$^{~\dagger, \mathsection}$, Amrutha Prasad$^{~\dagger, \mathparagraph}$, Seyyed Saeed Sarfjoo$^{~\dagger}$, Petr Motlicek$^{~\dagger}$
\thanks{The work was supported by the European Union's Horizon 2020 project No. 864702 - ATCO2 (Automatic collection and processing of voice data from air-traffic communications), which is a part of Clean Sky Joint Undertaking. The work was also partially supported by SESAR Joint Undertaking under Grant Agreement No. 884287, under European Union's Horizon 2020 Research and Innovation programme.}}
\address{
  $^{\dagger}$ Idiap Research Institute, Martigny, Switzerland \\
  $^{\ddagger}$ Institute of Computational Linguistics, University of Zürich \\
  $^{\mathsection}$ Ecole Polytechnique Federale de Lausanne (EPFL), Switzerland \\
  $^{\mathparagraph}$ Brno University of Technology, Brno, Czech Republic
}

\begin{document}
\ninept
\maketitle

\begin{abstract}

Automatic Speech Recognition (ASR), as the assistance of speech communication between pilots and air-traffic controllers, can significantly reduce the complexity of the task and increase the reliability of transmitted information.
ASR application can lead to a lower number of incidents caused by misunderstanding and improve air traffic management (ATM) efficiency.
Evidently, high accuracy predictions, especially, of key information, i.e., callsigns and commands, are required to minimize the risk of errors. We prove that combining the benefits of ASR and Natural Language Processing (NLP) methods to make use of surveillance data (i.e. additional modality)  helps to considerably improve the recognition of callsigns (named entity). In this paper, we investigate a two-step callsign boosting approach: (1)~at the \nth{1} step (ASR), weights of probable callsign n-grams are reduced in G.fst and/or in the decoding FST (lattices), (2)~at the \nth{2} step (NLP), callsigns extracted from the improved recognition outputs with Named Entity Recognition (NER) are correlated with the surveillance data to select the most suitable one. Boosting callsign n-grams with the combination of ASR and NLP methods eventually leads up to 53.7\% of an absolute, or 60.4\% of a relative, improvement in callsign recognition.
\end{abstract}

%




\section{Introduction}

Key components of speech communication between pilots and Air-Traffic Controllers (ATCo), i.e., callsigns, which are used for identification of aircrafts, and providing commands, demand high recognition accuracies. Callsigns are unique identifiers for aircrafts, of which the first part is an abbreviation of airline name and the last part is a flight number that contains a digit combination and may also incorporate an additional character combination, e.g., \textit{TVS84J} (see Table~\ref{tab:callsigns}). At a certain time point, only few aircrafts are usually in the radar zone which means only a limited number of callsigns can be referred to in the ATCo communications. If a recognized callsign does not match any `active' callsign registered by radar at the given time point, it means that there is no corresponding aircraft in the air space and the automatically recognized command (from voice communication) is invalid. Therefore, contextual information coming from the surveillance (radar) data allows adjusting system predictions that can significantly increase its accuracy.

\begin{table}[t]
  \caption{Callsigns: compressed and extended (airlines designators are in bold)}
  \label{tab:callsigns}
  \centering
  \begin{tabular}{ ll }
    \toprule
    Callsign & Extended callsign \\
    \midrule
    \textbf{SWR}2689 & \textbf{swiss} two six eight nine \\
    \textbf{RYR}1RK & \textbf{ryanair} one romeo kilo \\
    \textbf{RYR}1SG & \textbf{ryanair} one sierra golf \\
    \bottomrule
  \end{tabular}
\end{table}

Although contextual information has been already used in previous ATC studies~\cite{schmidt2014context,shore2012knowledge,oualil2015real,oualil2017context}, or more recently in~\cite{kocour21_interspeech,Iuliia_SUBMITTEDTOINTERSPEECH2021_2021,zuluagagomez21_interspeech}; it has been never adapted for both ASR and concept extraction outputs simultaneously and without a need of any additional knowledge (e.g., manual annotation, classes, etc.). This research aims to leverage the available contextual information by combining ASR and NLP methods. We believe that ASR and NLP are complementary tasks rather than separated ones. Whereas ASR exploits speech to produce a sequence of words, NLP exploits the intrinsic characteristics in a given snippet of text. ASR normally struggles to model long sequences, while state-of-the-art NLP systems allow extracting key information in the whole chunks of text; for instance an entire ATC utterance. In the proposed approach, we focus on an iterative use of contextual data, to take advantage of a combination of ASR and NLP modules. (1)~First, boosting the probability of active callsigns in ASR system (\textit{FST-boosting}), (2)~second, boosting ASR outputs (\textit{NLP-boosting}) in order to correct those predicted callsigns, which are not present in the surveillance data.


The rest of the paper is organised as follows: Section 2 reviews current approaches on integrating contextual knowledge in ASR for ATC communications. Section 3 gives a theoretical background of the proposed ASR-NLP approach to leverage surveillance data. Then, we present the data and the experiment set up in Section 4. Finally, we report the results and summarise our observations and ideas in Section 5 and 6, respectively.

\section{Contextual information for callsign detection}
Contextual data on the ASR level can be integrated by modifying weights of target n-grams in the grammar or/and in the ASR decoding lattices, e.g. by mean of generalised composition of baseline LM and Weighted Finite State Transducers (WFSTs) with the target contextual n-grams~\cite{hall2015composition,aleksic2015bringing,serrino2019contextual}.
A similar approach has been recently adopted in the ATC domain~\cite{kocour21_interspeech,Iuliia_SUBMITTEDTOINTERSPEECH2021_2021} and proved to offer a significant gain in callsign recognition. A list of callsigns to be boosted is regularly changing and needs to be updated dynamically per each utterance. Thus, weights of callsign n-grams are dynamically modified in the WFST. The first of the methods is lattice rescoring, where the weights are adjusted on the word recognition lattices from the first pass decoding. In the other method, weights are dynamically modified directly in the grammar (G.fst), which allows having target n-grams boosted before the decoding is performed~\cite{Iuliia_SUBMITTEDTOINTERSPEECH2021_2021}.
For our experiments, we will adopt the lattice rescoring approach to leverage the performance on the ASR side. 

Besides the ASR performance, contextual information for ATC has been also used to improve concept extraction~\cite{schmidt2014context,shore2012knowledge,oualil2015real,oualil2017context}. Schmidt~et~al.~\cite{schmidt2014context} applied a Context-Free Grammar (CFG)-based LM limiting the search space according to the contextual data. Shore~et~al.~\cite{shore2012knowledge} and Oualil~et~al.~\cite{oualil2015real,oualil2017context} build a CFG-based concept extractor with all semantic concepts of ATC embedded in XML annotation tags. In~\cite{shore2012knowledge}, after decoding, the lattice hypotheses are rescored by incorporating an additional knowledge source component to the cost function. The knowledge-based rescoring penalises hypotheses that are invalid in the context, e.g., callsigns not registered in the air space. In~\cite{oualil2015real}, to overcome the problem of variability of ATCO commands, the weighted Levenshtein distance is applied to find the closest match between an ASR hypothesis and generated context word sequences. \cite{oualil2017context} combines methods from~\cite{shore2012knowledge,oualil2015real} adding more contextual constraints from data with temporal information. Although these methods help to considerably increase the recognition accuracy, their limitation is that it deals only with concepts and callsigns which are annotated and included into the grammar. Those n-grams that do not appear in the grammar can not be extracted and evaluated. Finally, Helmke et al.~\cite{helmke2020machine} recently proposed a machine learning algorithm for command extraction from the ASR hypothesized outputs with the use of keywords. This model achieves good results and it is the second alternative approach to our methods.


\section{Methods}
\label{sec:methods}
\begin{figure}[t]
  \centering
  \includegraphics[width=\linewidth]{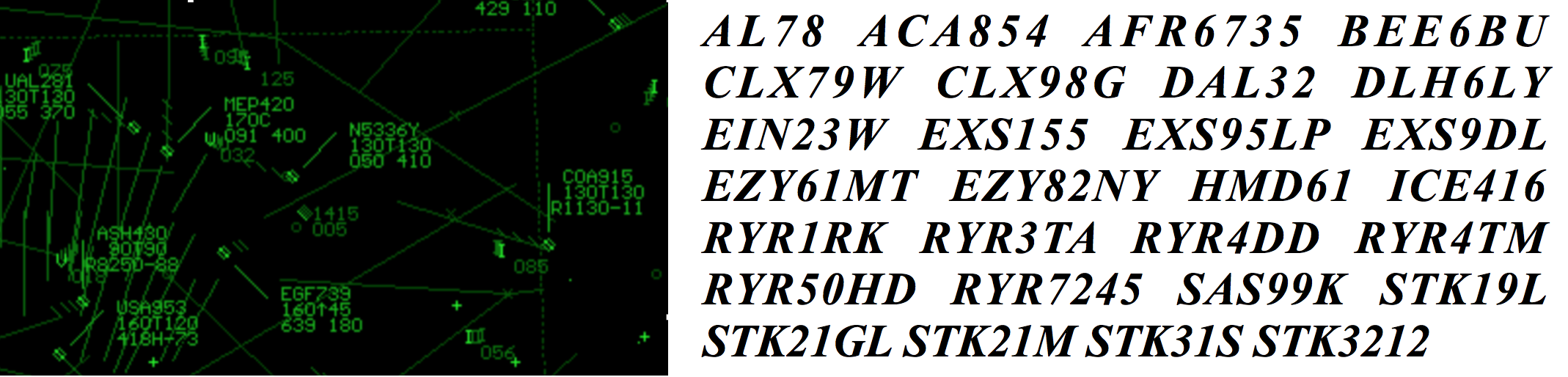}
  \caption{Callsigns in ICAO format received from radar.}
  \label{fig:pic_radar}
\end{figure}

We focus on the combination of ASR and NLP methods and investigate two-steps approach for callsigns extraction. As a callsign is a sequence of words, using contextual information to improve recognition of callsigns is a task of boosting n-grams. The contextual data comes from radar in a compressed form, i.e., standardized phraseology format of International Civil Aviation Organization
(ICAO)~\cite{icao} (see Fig.~\ref{fig:pic_radar}). To introduce the contextual knowledge into the ASR system, all callsigns need to be expanded to word sequences (Table~\ref{tab:callsigns}). The compressed form often allows more than one possible realisation in the ATCos' speech: For example, \textbf{DLH5KX} can be expanded as \textit{`hansa five kilo x-ray'} or \textit{`lufthansa five kilo x-ray'}, etc. As we can not say which particular expansion is true for an uttered callsign, it is important to take all expansion variants into account.

\subsection{Integration of contextual knowledge into ASR system}

In a standard hybrid-based ASR system, the different knowledge sources are represented as WFSTs, which are combined by the `composition' operator together in the final decoding graph~\cite{mohri2002weighted}. Information from additional knowledge sources can be also integrated into a system by means of composition.

Our first integration of contextual knowledge into ASR is done on the LM level (\textit{G-extension}). The idea is to boost callsign n-grams already available in LM, and even more important to add those callsign n-grams, which are absent (e.g., $>$3 words sequences in 3-gram LM). We build a contextual \textit{FST} that includes all possible callsigns from the tower: all callsigns registered by the radar at different time stamps (from 17K to 280K callsigns to boost in different test sets; see last column in Table~\ref{tab:test_sets}). Then, the main $G.fst$ is composed with the contextual $G\_biased.fst$ and the result of composition is used in the final decoding $HCLG$ graph.

The second integration of contextual information (\textit{lattice rescoring}) is done per utterance on top of the decoding lattices which allows flexible adaptation to new-coming contextual information avoiding changing the main decoding graph ($HCLG$) (for more details check~\cite{Iuliia_SUBMITTEDTOINTERSPEECH2021_2021}).
Weights in lattices are rescored according to the surveillance data: for each test utterance, an $FST$ biased to callsigns n-grams registered at the time stamp when an utterance is created and composed with lattices created in the first pass:
%
\begin{equation}
  Lattices' = Lattices \circ biasing\_FST
  \label{fst-biased-composition}
\end{equation}
Weights updated in the composition are used for final predictions.




\subsection{Integration of contextual knowledge on ASR transcripts}

Our approach for integrating contextual knowledge on ASR transcripts (e.g., 1-best hypothesis) is based on a two-step pipeline. Each step conveys an independent module.

\subsubsection{Named Entity Recognition (NER) module}

\begin{figure}[t]
  \centering
  \includegraphics[width=\linewidth]{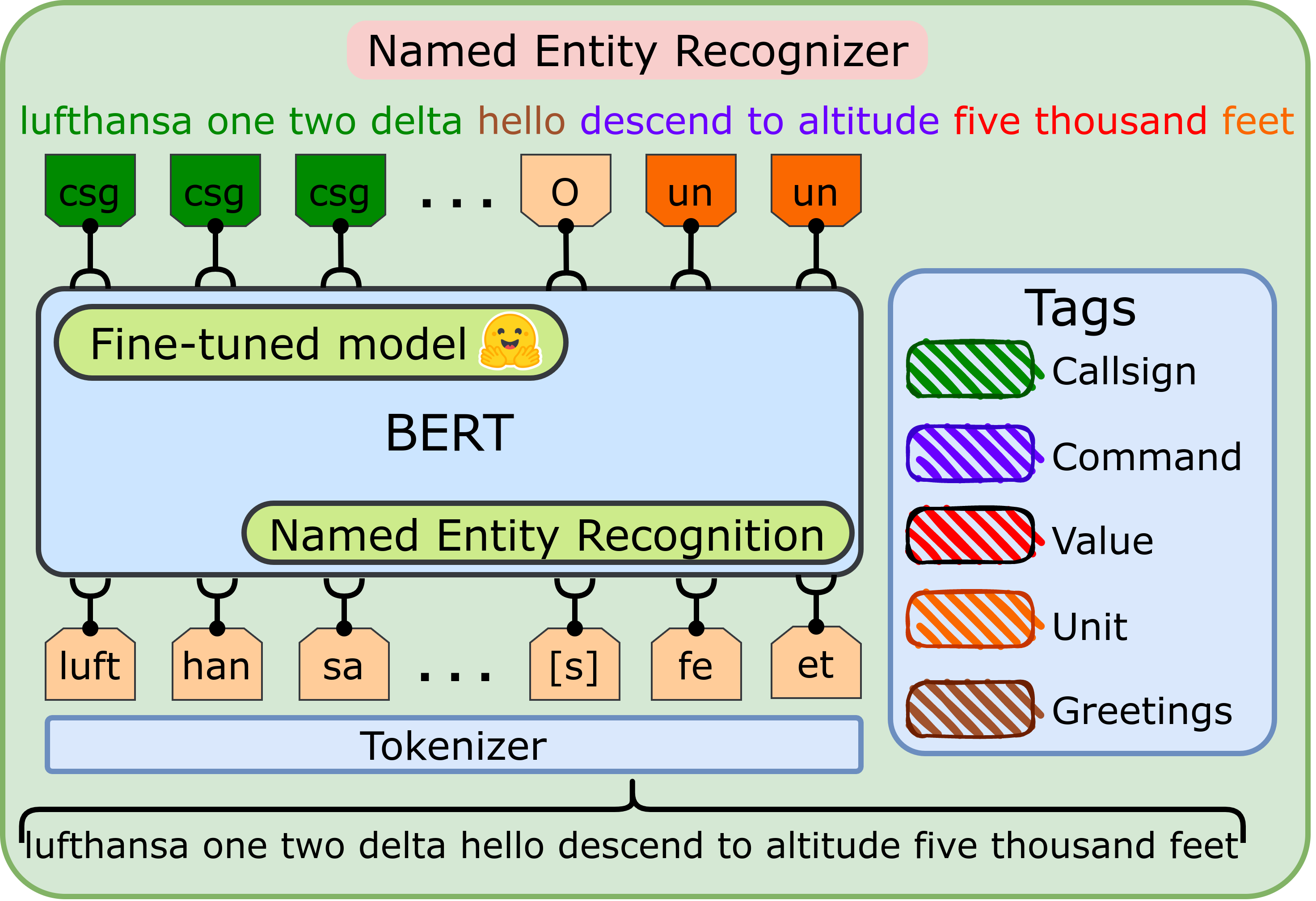}
  \caption{BERT-based model (Huggingface) fine-tuned on NER task.}
  \label{fig:ner_pipeline}
\end{figure}

ATC communications carry rich information such as callsigns, commands, values and units; they can be seen as `named entities'. We propose a NLP-based system to extract such information from ASR transcripts. We defined callsigns, commands, units, values, greetings OR the rest (e.g., `None' class) as tags for the NER task, as depicted in Figure~\ref{fig:ner_pipeline}. First, we downloaded a BERT~\cite{devlin2018bert} model pre-trained as masked language model from Huggingface~\cite{wolf-etal-2020-transformers} and fine-tuned it on NER task with 12k sentences ($\sim$12 hours of speech), where each word has a tag. Then, we developed a data augmentation pipeline in order to increase the amount of training data: 1M samples from 12k sentences. The pipeline has four actions that modifies the training sample: \textit{add}, \textit{delete}, \textit{swap}, or \textit{move} the \textbf{callsign} across the utterance -sentence-. \textit{Delete} and \textit{move} actions, remove and keep the same callsigns, respectively; \textit{add} and \textit{swap} generate a sentence with a new callsign picked randomly from a callsign list. The callsign list is pre-defined by a user, which makes the approach easy to deploy in out-of-domain data (i.e., callsigns from different airports/countries).


\subsubsection{Re-ranking module based on Levenshtein distance}
\label{sec:reranking}

The BERT-based system for NER allows us to extract the callsign from a given transcript or ASR 1-best hypotheses. Recognition of this entity is crucial where a single error produced by the ASR system affects the whole entity (normally composed of three to eight words). Additionally, speakers regularly shorten callsigns in the conversation making it impossible for an ASR system to generate the full entity (e.g., \textit{`three nine two papa'} instead of \textit{`austrian three nine two papa'}, \textit{`six lima yankee'} instead of \textit{`hansa six lima yankee'}). One way to overcome this issue is to re-rank entities extracted by the BERT-based NER system with the surveillance data. The output of an NER system is a list of tags that match words or sequences of words in an input utterance. As our only available source of contextual knowledge are callsigns registered at a certain time and location, we extract callsigns with the NER system and discard other entities. Correspondingly, each utterance has a list of callsigns expanded into word sequences (shown in Table~\ref{tab:callsigns}). As input, the re-ranking module takes (i)~a callsign extracted by the NER system and (ii)~an expanded list of callsigns. The re-ranking module compares a given n-gram sequence against a list of possible n-grams, and finds the closest match from the list of surveillance data based on the weighted Levenshtein distance.
We skip the re-ranking in case the NER system outputs a `NO\_CALLSIGN' flag (no callsign recognized).

\section{Data and experimental setup}

\subsection{Data}


For the callsign boosting experiments, we use four test sets; all of them have utterances both with and without callsigns (see Table~\ref{tab:test_sets}). 



\textbf{LiveATC:} the first test set is from the LiveATC\footnote{Streaming audio platform that gathers VHF aircraft communications} data recorded from publicly accessible VHF radio channels, which includes both pilots and ATCo speech and, therefore, is of rather low quality (i.e., low SNR often below 10dB)~\cite{zuluaga2020callsign}.

\textbf{MALORCA:} Prague and Vienna test sets are mainly of good quality (i.e., telephone quality speech with SNR usually above 20dB) data from the MALORCA project~\cite{Khonglah_ICASSP2020_2020, srinivasamurthy2017semi}\footnote{From the `standard' MALORCA test sets~\cite{srinivasamurthy2017semi} only utterances with the available surveillance information are selected.} which includes only ATCo speech. The recognition accuracy of the baseline model are already high above the one reached on VHF LiveATC data (see Table~\ref{tab:results_nlp_boosting}). The data was collected from the Prague and Vienna airports and, thus, forms two separate sets correspondingly.

\textbf{NATS:} a data set collected under HAAWAII project\footnote{\url{https://www.haawaii.de/wp/}} with the data coming from London approach (airport).
This data is relatively high-quality, similar to MALORCA. 

The data sets are used differently in training ASR and NER models. The ASR train data includes Malorca sets but not LiveATC and NATS. The data for fine-tuning the NER system contains LiveATC data but neither Malorca, nor NATS sets. 

\begin{table}[t]
  \caption{Test sets (callsigns (csgn) per utterance (utt) \textemdash{} median of callsigns per utterance in the surveillance data)}
  \label{tab:test_sets}
  \centering
  \resizebox{0.42\textwidth}{!}{
      \begin{tabular}{ lccccc }
    \toprule
    & \multicolumn{2}{c}{N of utt} & Csgn& & \\
    Test set & with & w/o & per & Min & All csgns \\
    & \multicolumn{2}{c}{a csgn} & utt & & \\
    \midrule
    LiveATC & 581 & 29 & 28 & 40 & 280K \\
    Malorca Prague & 784 & 88 & 5 & 82 & 17K \\
    Malorca Vienna & 877 & 38 & 19 & 65 & 59K \\
    NATS & 794 & 73 & 50 & 50 & 168K \\
    \bottomrule
  \end{tabular}
  }
\end{table}

\subsection{ASR model}
\label{subsec:model}

\begin{table*}[t]
  \caption{Results of callsign extraction with ASR boosting (ASR-B) and post-boosting (NLP-B): the accuracy of callsign recognition (\%) is calculated for the callsigns in ICAO format (see Section~\ref{sec:evaluation})}
  \label{tab:results_nlp_boosting}
  \centering
      \begin{tabular}{ llcccccc }
    \toprule
    \textbf{Method} & & & & \multicolumn{4}{c}{\textbf{Test sets (callsign recognition accuracy)}} \\
    \cline{5-8}
    \rule{0pt}{3ex} & & & & LiveATC & Prague & Vienna & NATS \\

    \cline{1-8}
    \textbf{ASR output} & ~~Callsign extraction (baseline) & & & 42.8 & 64.4 & 48.4 & 35.2 \\
    & \quad ~~~~~~~~~~~~~~~~~~~~Lattice rescoring & G-extension & NLP-boosting & & & & \\
    & \quad ~~~~~~~~~~~~~~~~~~~~~~~~~~~~~~~~\checkmark & - & - & 53.1 & 66.9 & 59.6 & 37.1 \\
    & \quad ~~~~~~~~~~~~~~~~~~~~~~~~~~~~~~~~- & \checkmark & - & 44.4 & 64.3 & 49.2 & 34.8 \\
    & \quad ~~~~~~~~~~~~~~~~~~~~~~~~~~~~~~~~\checkmark & \checkmark & - & 52.8 & 66.9 & 52.1 & 36.8 \\
    & \quad ~~~~~~~~~~~~~~~~~~~~~~~~~~~~~~~~- & - & \checkmark & 88.4 & \textbf{95.0} & \textbf{86.0} & 87.0 \\
    & \quad ~~~~~~~~~~~~~~~~~~~~~~~~~~~~~~~~\checkmark & - & \checkmark & \textbf{88.5} & 94.8 & 84.3 & \textbf{88.9} \\
    & \quad ~~~~~~~~~~~~~~~~~~~~~~~~~~~~~~~~- & \checkmark & \checkmark & 87.7 & \textbf{95.0} & 85.6 & 88.2 \\
    & \quad ~~~~~~~~~~~~~~~~~~~~~~~~~~~~~~~~\checkmark & \checkmark & \checkmark & 88.0 & 94.7 & 84.0 & 88.0 \\
    \midrule
    \textbf{Ground Truth} & ~~Callsign extraction (oracle) & & & \textbf{89.7} & 72.2 & 59.6 & 67.4 \\
    & \quad + NLP-Boosting & & & 89.3 & \textbf{95.4} & \textbf{87.0} & \textbf{94.0} \\
    \bottomrule
    \textbf{ASR WER} & \textbf{(without boosting)} & & & 32.4 & 3.4 & 9.2 & 24.4 \\
    \bottomrule
  \end{tabular}
\end{table*}

For training the baseline acoustic model, as well as for the decoding and rescoring experiments, we used the Kaldi framework~\cite{povey2011kaldi}. The system follows the standard Kaldi recipe, which uses MFCC and i-vectors features. The standard chain training is based on Lattice-free MMI (LF-MMI)~\cite{povey2016purely}, which includes 3-fold speed perturbation and one third frame sub-sampling.

The acoustic model is a CNN-TDNNF trained on approximately 1200 hours of ATC labeled augmented data~\cite{zuluagagomez20_interspeech,zuluaga2020callsign}. First, the training databases (195~hours\footnote{The ATCO2 test set is publicly available in \url{https://www.atco2.org/data}}) were augmented by adding noises that match LiveATC audio channel (one batch between 5-10 dB and other 10-20dB SNR). Afterwards, we applied speed perturbation, obtaining almost 1200 hours of training data. The model was further improved with 700~hours of semi-supervised data collected in LiveATC for different airports from Europe~\cite{Khonglah_ICASSP2020_2020}.
The LM is 3-gram trained on the same data as the acoustic model with an additional textual data from additional public resources such as airlines names, airports, ICAO alphabet and way-points in Europe.


\subsection{Evaluation}
\label{sec:evaluation}
Since this paper focuses on improving callsign detection, we evaluate the proposed methods by calculating the accuracy of callsign extraction. For the evaluation we use ICAO format, which is the target form to display on the screen of ATCo and pilots, and we have only two outcomes: ICAO is recognized `correctly' VS `incorrectly'. In the previous studies~\cite{kocour21_interspeech,Iuliia_SUBMITTEDTOINTERSPEECH2021_2021}, the accuracy of callsign recognition is evaluated with matching the ground truth callsign n-grams to the ones in utterances. This approach, however, does not correspond to the real situation, when ground truth callsigns are not available. In our experiments, we do not only do speech recognition but proceed with callsign extraction, we evaluate the performance directly on the extracted entities. In addition, the use of the ICAO format helps to avoid issues with variability of pronunciation within a callsign: the full form of callsign is extracted automatically but a speaker says a shorten version, which is then outputted by the ASR, as well as recorded in the ground truth transcriptions (see example above~\ref{sec:reranking}).
All experiments share the same ASR and BERT-based NER systems, as well as the ICAO extractor module; thus, the performances are only impacted by the proposed boosting techniques.

\section{Results}
As a baseline we use callsign extraction done directly on the outputs of our ASR system. Then, we apply the proposed boosting techniques (G-extension, lattice rescoring, NLP-boosting) in different combinations to see how they can benefit from each other. In Table~\ref{tab:results_nlp_boosting}, the results of the experiments are presented on four different test sets with accuracy of callsign (ICAO) recognition. Overall, the proposed metrics help to improve the baseline accuracy from 30.6\% to 53.7\% absolutely, or from 32.1\% to 60.4\% relatively (for the test sets Prague and NATS correspondingly; when the NATS set gets the highest improvement being the out-of-domain data). The best results are always achieved with the use of NLP-boosting. For LiveATC and NATS sets, the out-of-domain sets in the ASR training, the best performance is achieved with the combination of NLP-boosting and ASR-boosting (lattice rescoring) methods.

At the same time, the G-extension has a contradicting effect. It helps to improve results comparing to the baseline for the LiveATC and Vienna sets, yet, its combination with lattice rescoring achieves worse accuracy than lattice rescoring alone.
The possible drawback of the G-extension method is that a very high number of available callsigns are boosted in LM $FST$ (see last column~\ref{tab:test_sets}). It can introduce confusion when combining with the lattice rescoring boosting method, which focuses on only current callsigns.
On the other hand, it does not need any modifications during the decoding and serves as a general domain adaptation. Thus, G-extension can be used to improve the outputs when other methods are not available, otherwise, can be skipped.
The number of callsigns used to boost the ASR outputs may also have the degradation effect on the performance of the lattice rescoring approach. Although in this case, the number of callsigns did not exceed 50, we investigated its impact. The test sets have different numbers of boosted n-grams, from 5 to 50 (see Table~\ref{tab:callsigns}), but even with 50 boosted callsigns the recognition accuracy goes considerably up comparing to the baseline.

Along with the evaluation of boosting methods on the ASR outputs, we provide the `oracle' results, when callsigns are extracted on the ground truth transcriptions (\nth{2} line in Table~\ref{tab:results_nlp_boosting}). This comparison allows estimating the impact of the proposed methods to the callsign extraction improvement, when no ground truth information is available. Even if the `oracle' scores always stay better, the accuracy achieved with our systems shows close and comparable results. No improvement with NLP-boosting on the ground truth transcription for LiveATC test set can be explained by already high accuracy of callsign extraction, as LiveATC data was used to fine-tune the NER.

Table~\ref{tab:improve_examples} gives examples of improvement where airline names and callsigns are detected correctly comparing to the baseline predictions. Our methods demonstrate consistent results for data of different quality. The level of noise in the recordings of LiveATC and Malorca test sets is very different, as well as WERs achieved by their baseline ASR systems (the last line in Table~\ref{tab:results_nlp_boosting}; \cite{Iuliia_SUBMITTEDTOINTERSPEECH2021_2021}). Nevertheless, we see considerable improvement for all test sets and the general tendency stays the same. The main advantage of the proposed approach comparing to the others is its simplicity and flexibility. The NER-system can be fine-tuned to different data sets that makes it easy to adapt to new out-of-domain data. Moreover, it is also suitable for the online implementation.

\begin{table}[t]
  \caption{Examples of improved callsign recognition (bold part)}
  \label{tab:improve_examples}
  \centering
  \resizebox{0.47\textwidth}{!}{
  \begin{tabular}{ p{4.1cm} p{3.6cm} }
    \toprule
    Baseline (incorrect ICAO) & Boosted (correct ICAO) \\
    \midrule
    \textbf{wizz air} four one six (\textbf{WZZ}416) & \textbf{iceair} four one six (\textbf{ICE}416) \\
    \textbf{easy} three delta (\textbf{EZY}3D) & \textbf{fraction eight eight} three delta (\textbf{NJE88}3D) \\
    \textbf{serbia} one nine lima (\textbf{ASL}19L) & \textbf{stobart} one nine lima (\textbf{STK}19L) \\
    \bottomrule
  \end{tabular}
  }
\end{table}

\section{Conclusion}

We investigated a two-step approach of integrating contextual radar data in order to dynamically improve the recognition of callsigns per utterance. We demonstrated that the best result is achieved with the NLP-boosting and with the combination of NLP-boosting and lattice rescoring methods on all test sets of different recording quality with the significant improvement, i.e., from 32.1\% to 60.4\% of relative improvement on callsign recognition accuracy across the evaluated data sets. Introduction of contextual information considerably improves recognition of callsigns and, thus, recognition of ATCo messages in general. As a noisy environment leading to lower recognition accuracy is often a reality in pilot-ATCo communication, the proposed methods and their combination will definitely benefit the recognition of the key information in ATCo speech.


\bibliographystyle{IEEEtran}

\bibliography{references}

\begin{thebibliography}{10}

\bibitem{schmidt2014context}
Anna Schmidt, Youssef Oualil, Oliver Ohneiser, Matthias Kleinert, Marc
  Schulder, Arif Khan, Hartmut Helmke, and Dietrich Klakow,
\newblock ``Context-based recognition network adaptation for improving on-line
  asr in air traffic control,''
\newblock in {\em 2014 IEEE Spoken Language Technology Workshop (SLT)}. IEEE,
  2014, pp. 13--18.

\bibitem{shore2012knowledge}
Todd Shore, Friedrich Faubel, Hartmut Helmke, and Dietrich Klakow,
\newblock ``Knowledge-based word lattice rescoring in a dynamic context,''
\newblock in {\em Thirteenth Annual Conference of the International Speech
  Communication Association}, 2012.

\bibitem{oualil2015real}
Youssef Oualil, Marc Schulder, Hartmut Helmke, Anna Schmidt, and Dietrich
  Klakow,
\newblock ``Real-time integration of dynamic context information for improving
  automatic speech recognition,''
\newblock in {\em Sixteenth Annual Conference of the International Speech
  Communication Association}, 2015.

\bibitem{oualil2017context}
Youssef Oualil, Dietrich Klakow, Gyorgy Szasz{\'a}k, Ajay Srinivasamurthy,
  Hartmut Helmke, and Petr Motlicek,
\newblock ``A context-aware speech recognition and understanding system for air
  traffic control domain,''
\newblock in {\em 2017 IEEE Automatic Speech Recognition and Understanding
  Workshop (ASRU)}. IEEE, 2017, pp. 404--408.

\bibitem{kocour21_interspeech}
Martin Kocour, Karel Vesel{\`y}, Alexander Blatt, Juan~Zuluaga Gomez, Igor
  Sz{\"o}ke, Jan Cernocky, Dietrich Klakow, and Petr Motlicek,
\newblock ``{Boosting of Contextual Information in ASR for Air-Traffic
  Call-Sign Recognition},''
\newblock in {\em Proc. Interspeech 2021}, 2021, pp. 3301--3305.

\bibitem{Iuliia_SUBMITTEDTOINTERSPEECH2021_2021}
Iuliia Nigmatulina, Rudolf Braun, Juan Zuluaga-Gomez, and Petr Motlicek,
\newblock ``Improving callsign recognition with air-surveillance data in
  air-traffic communication,''
\newblock Idiap Research Institute, 2021, pp. 1--5, Idiap Research Institute.

\bibitem{zuluagagomez21_interspeech}
Juan Zuluaga-Gomez, Iuliia Nigmatulina, Amrutha Prasad, Petr Motlicek, Karel
  Vesel{\`y}, Martin Kocour, and Igor Sz{\"o}ke,
\newblock ``{Contextual Semi-Supervised Learning: An Approach to Leverage
  Air-Surveillance and Untranscribed ATC Data in ASR Systems},''
\newblock in {\em Proc. Interspeech 2021}, 2021, pp. 3296--3300.

\bibitem{hall2015composition}
Keith Hall, Eunjoon Cho, Cyril Allauzen, Francoise Beaufays, Noah Coccaro,
  Kaisuke Nakajima, Michael Riley, Brian Roark, David Rybach, and Linda Zhang,
\newblock ``Composition-based on-the-fly rescoring for salient n-gram
  biasing,''
\newblock 2015.

\bibitem{aleksic2015bringing}
Petar Aleksic, Mohammadreza Ghodsi, Assaf Michaely, Cyril Allauzen, Keith Hall,
  Brian Roark, David Rybach, and Pedro Moreno,
\newblock ``Bringing contextual information to google speech recognition,''
\newblock 2015.

\bibitem{serrino2019contextual}
Jack Serrino, Leonid Velikovich, Petar~S Aleksic, and Cyril Allauzen,
\newblock ``Contextual recovery of out-of-lattice named entities in automatic
  speech recognition.,''
\newblock in {\em Interspeech}, 2019, pp. 3830--3834.

\bibitem{helmke2020machine}
Hartmut Helmke, Matthias Kleinert, Oliver Ohneiser, Heiko Ehr, and Shruthi
  Shetty,
\newblock ``Machine learning of air traffic controller command extraction
  models for speech recognition applications,''
\newblock in {\em 2020 AIAA/IEEE 39th Digital Avionics Systems Conference
  (DASC)}. IEEE, 2020, pp. 1--9.

\bibitem{icao}
``All clear phraseology manual,''
\newblock in {\em Eurocontrol, Brussels, Belgium}, 2011,
\newblock "[Online; accessed 10-September-2021]".

\bibitem{mohri2002weighted}
Mehryar Mohri, Fernando Pereira, and Michael Riley,
\newblock ``Weighted finite-state transducers in speech recognition,''
\newblock {\em Computer Speech \& Language}, vol. 16, no. 1, pp. 69--88, 2002.

\bibitem{devlin2018bert}
Jacob Devlin, Ming-Wei Chang, Kenton Lee, and Kristina Toutanova,
\newblock ``Bert: Pre-training of deep bidirectional transformers for language
  understanding,''
\newblock {\em arXiv preprint arXiv:1810.04805}, 2018.

\bibitem{wolf-etal-2020-transformers}
Thomas Wolf, Lysandre Debut, Victor Sanh, Julien Chaumond, Clement Delangue,
  Anthony Moi, Pierric Cistac, Tim Rault, Rémi Louf, and Morgan~Funtowicz
  et~al,
\newblock ``Transformers: State-of-the-art natural language processing,''
\newblock in {\em Proceedings of the 2020 Conference on Empirical Methods in
  Natural Language Processing: System Demonstrations}. 2020, pp. 38--45,
  Association for Computational Linguistics.

\bibitem{zuluaga2020callsign}
Juan Zuluaga-Gomez, Karel Vesel{\`y}, Alexander Blatt, Petr Motlicek, Dietrich
  Klakow, Allan Tart, Igor Sz{\"o}ke, Amrutha Prasad, Saeed Sarfjoo, Pavel
  Kol{\v{c}}{\'a}rek, et~al.,
\newblock ``Automatic call sign detection: Matching air surveillance data with
  air traffic spoken communications,''
\newblock in {\em Multidisciplinary Digital Publishing Institute Proceedings},
  2020, vol.~59, p.~14.

\bibitem{Khonglah_ICASSP2020_2020}
Banriskhem Khonglah, Srikanth Madikeri, Subhadeep Dey, Herv{\'{e}} Bourlard,
  Petr Motlicek, and Jayadev Billa,
\newblock ``Incremental semi-supervised learning for multi-genre speech
  recognition,''
\newblock in {\em Proceedings of ICASSP 2020}, 2020.

\bibitem{srinivasamurthy2017semi}
Ajay Srinivasamurthy, Petr Motlicek, Ivan Himawan, Gyorgy Szaszak, Youssef
  Oualil, and Hartmut Helmke,
\newblock ``Semi-supervised learning with semantic knowledge extraction for
  improved speech recognition in air traffic control,''
\newblock in {\em Proc. of the 18th Annual Conference of the International
  Speech Communication Association}, 2017.

\bibitem{povey2011kaldi}
Daniel Povey, Arnab Ghoshal, Gilles Boulianne, Lukas Burget, Ondrej Glembek,
  Nagendra Goel, Mirko Hannemann, Petr Motlicek, Yanmin Qian, Petr Schwarz,
  et~al.,
\newblock ``The kaldi speech recognition toolkit,''
\newblock in {\em IEEE workshop on automatic speech recognition and
  understanding}. IEEE Signal Processing Society, 2011.

\bibitem{povey2016purely}
Daniel Povey, Vijayaditya Peddinti, Daniel Galvez, Pegah Ghahremani, Vimal
  Manohar, Xingyu Na, Yiming Wang, and Sanjeev Khudanpur,
\newblock ``Purely sequence-trained neural networks for asr based on
  lattice-free mmi.,''
\newblock in {\em Interspeech}, 2016, pp. 2751--2755.

\bibitem{zuluagagomez20_interspeech}
Juan Zuluaga-Gomez, Petr Motlicek, Qingran Zhan, Karel Vesel{\`y}, and Rudolf
  Braun,
\newblock ``{Automatic Speech Recognition Benchmark for Air-Traffic
  Communications},''
\newblock in {\em Proc. Interspeech 2020}, 2020, pp. 2297--2301.

\end{thebibliography}


\end{document}